\documentclass[letterpaper, 10 pt, journal, twoside]{IEEEtran}

\usepackage[utf8]{inputenc}

\usepackage{mathbbol}

\usepackage{esvect}

\usepackage[percent]{overpic}

\usepackage{tabu}

\usepackage{multirow}

\def\mc{\multicolumn}

\usepackage[overload]{empheq}

\usepackage{bm}

\usepackage{xcolor}
\definecolor{dg}{rgb}{0.1, 0.6, 0.2}       
\definecolor{b}{rgb}{0.0, 0.0, 1}          

\usepackage{epsfig}
\usepackage{float}
\usepackage{color}
\usepackage{url}

\usepackage{algpseudocode}
\usepackage[ruled,linesnumbered,vlined]{algorithm2e}

\usepackage{amssymb, amsfonts}
\usepackage{amsmath}
\usepackage{mathrsfs}

\usepackage{amsthm}
\usepackage{mathtools}

\let\labelindent\relax
\usepackage{enumitem}       
\setenumerate[enumerate]{align=left}

\usepackage{graphicx}
\usepackage{subcaption}
\usepackage{caption}

\usepackage{cite}

\makeatletter
\let\NAT@parse\undefined
\makeatother
\usepackage{hyperref} 

\usepackage{balance}

\newcommand{\norm}[1]{\left\lVert#1\right\rVert}
\newcommand{\abs}[1]{\left\lvert#1\right\rvert}

\makeatletter
\newlength\tmp@\newlength\t@mp
\newcommand{\comp}[3]
  {\mathop{ \settowidth\tmp@{$\displaystyle\mathop{#1}^{#3}_{#2}$}
  \hbox to \tmp@{\hss \settowidth\t@mp{$\displaystyle #1$}\setlength\t@mp{.45\t@mp}
  $\displaystyle\mathop{#1}^{\hspace\t@mp #3}_{\hspace{-\t@mp}#2}$
  \hss} }}
\makeatother

\newcommand{\Int}[2]
{\int_{#1}^{#2}}

\DeclareMathOperator*{\argmin}{argmin}

\hypersetup{
    colorlinks=true,
    linkcolor=blue,
    filecolor=blue,      
    urlcolor=blue,
    citecolor=blue,
}

\def\a{\alpha}
\def\b{\beta}

\def\g{\gamma}

\def\o{\omega}

\def\D{\Delta}

\def\R{\mathbb{R}}

\def\l{\left}
\def\r{\right}

\def\quat{\mathbf{q}}

\def\pos{\mathbf{p}}
\def\vel{\mathbf{v}}
\def\accel{\mathbf{a}}

\newcommand{\fr}[1]{\texttt{#1}}

\newcommand{\vbf}[1]{{\bm{\mathbf{#1}}}}

\def\calE{\mathcal{E}}

\def\resi{\bm{r}}

\def\bias{\mathbf{b}}

\def\rot{\mathbf{R}}
\def\tf{\mathbf{T}}

\def\SO{\mathrm{SO(3)}}

\def\SE{\mathrm{SE(3)}}

\def\qtoR{\mathcal{R}}

\def\Dt{ {\D t} }

\def\wrt{\text{w.r.t. }}

\def\pia{{\vbf{\a}}}
\def\pib{{\vbf{\b}}}
\def\pig{{\vbf{\g}}}

\def\X{\mathcal{X}}

\def\I{\mathcal{I}}

\def\M{\mathcal{M}}
\def\L{\mathcal{L}}
\def\P{\mathcal{P}}
\def\F{\mathcal{F}}
\def\C{\mathcal{C}}

\def\angvel{\bm{\omega}}
\def\accel{\mathbf{a}}
\def\grav{\mathbf{g}}

\newtheorem{remark}{Remark}

\def\fA{\fr{A}}
\def\fB{\fr{B}}

\def\fL{\fr{L}}

\def\Xhat{\hat{\X}}

\def\f{\vbf{f}}
\def\n{\vbf{n}}
\def\Fbrv{\breve{\F}}

\def\Q{\mathbb{Q}}
\def\q{\mathbb{q}}

\newcommand{\ul}[1]{\underline{#1}}

\newcommand{\highlight}[1]{{ #1}}

\begin{document}

\onecolumn
\thispagestyle{empty}

\hspace{3cm}
\begin{center}
This paper has been accepted for publication in \emph{IEEE Robotics And Automation Letters} (RA-L).\\

\hspace{1cm}

DOI: \href{http://dx.doi.org/10.1109/LRA.2021.3080633}{10.1109/LRA.2021.3080633}\\

IEEE Explore: \url{https://ieeexplore.ieee.org/document/9431754} \\

\hspace{1cm}

Please cite the paper using the following bibtex code: \\

\hspace{1cm}

\begin{verbatim}

@article{nguyen2021miliom,
  title={MILIOM: Tightly Coupled Multi-Input Lidar-Inertia Odometry and Mapping},
  author={Nguyen, Thien-Minh
          and Yuan, Shenghai
          and Cao, Muqing
          and Yang, Lyu
          and Nguyen, Thien Hoang
          and Xie, Lihua},
  journal={IEEE Robotics and Automation Letters},
  volume={6},
  number={3},
  pages={5573--5580},
  year={2021},
  publisher={IEEE}
}

\end{verbatim}

\end{center}
\twocolumn

\setcounter{page}{1}

\title{MILIOM: Tightly Coupled Multi-Input Lidar-Inertia 
Odometry and Mapping}
\author{Thien-Minh Nguyen$^{1}$, \IEEEmembership{Member,~IEEE},
        Shenghai Yuan$^{1}$,
        Muqing Cao$^{1}$,
        Lyu Yang$^{1}$,\\
        Thien Hoang Nguyen$^{1}$, \IEEEmembership{Student Member,~IEEE}
		and Lihua Xie$^{1}$, \IEEEmembership{Fellow,~IEEE}%
\thanks{Manuscript received: February 24, 2021; Revised April 21, 2021; Accepted May 12, 2021.}%
\thanks{This paper was recommended for publication by Editor Javier Civera upon evaluation of the Associate Editor and Reviewers' comments.
This work was supported by the Wallenberg AI, Autonomous Systems and Software Program (WASP) funded by the Knut and Alice Wallenberg Foundation, under the Grant Call 10013 - Wallenberg-NTU Presidential Postdoctoral Fellowship 2020. (Corresponding Author: Thien-Minh Nguyen)}%
\thanks{$^{1}$The authors are with School of Electrical and Electronic Engineering, Nanyang Technological University, Singapore 639798, 50 Nanyang Avenue. (e-mail:
{\tt\footnotesize \{thienminh.nguyen@, shyuan@, mqcao@, lyu.yang@, e180071@e., elhxie@\}ntu.edu.sg}).}%
\thanks{Digital Object Identifier (DOI): see top of this page.}
}

\markboth{IEEE Robotics and Automation Letters. Preprint Version. Accepted May, 2021}
{Nguyen \MakeLowercase{\textit{et al.}}: MILIOM} 

\maketitle

\begin{abstract}
In this paper we investigate a tightly coupled Lidar-Inertia Odometry and Mapping (LIOM) scheme, with the capability to incorporate multiple lidars with complementary field of view (FOV). In essence, we devise a time-synchronized scheme to combine extracted features from separate lidars into a single pointcloud, which is then used to construct a local map and compute the feature-map matching (FMM) coefficients. These coefficients, along with the IMU preinteration observations, are then used to construct a factor graph that will be optimized to produce an estimate of the sliding window trajectory. We also propose a key frame-based map management strategy to marginalize certain poses and pointclouds in the sliding window to grow a global map, which is used to assemble the local map in the later stage. The use of multiple lidars with complementary FOV and the global map ensures that our estimate has low drift and can sustain good localization in situations where single lidar use gives poor result, or even fails to work. Multi-thread computation implementations are also adopted to
fractionally cut down the computation time and ensure real-time performance. We demonstrate the efficacy of our system via a series of experiments on public datasets collected from an aerial vehicle.

\end{abstract}


\IEEEpeerreviewmaketitle

\section{Introduction}

Over the years, 3D lidars have proved to be a reliable and accurate localization solution for autonomous systems. Compared with the other class of onboard self-localization (OSL) based on camera, lidar clearly possesses many superior characteristics. In terms of sensing capability, lidar can provide 360 degree observation of the environment, with metric-scaled features, while the best fisheye cameras still only have up to $180^0$ field of view (FOV), and can only extract features of unknown scale. Even when compared to multi-camera or RGBD camera systems, which are supposed to solve the scale problem, lidar still has a much higher sensing range, is almost unaffected by lighting conditions, and requires very little calibration effort.
Nevertheless, high cost and weight has been one of lidar's main weaknesses, and has mostly restricted its application to ground robot platforms. However, in recent years, both the cost and weight factors have been significantly reduced thanks to many technological innovations. Hence, we have observed an uptick of 3D lidar applications in the robotics community.

Recently some researchers have started investigating the use of multiple 3D lidars for better coverage of the environment, especially in the field of autonomous driving \cite{jeong2019complex, palieri2020locus, jiao2020robust, geyer2020a2d2, agarwal2020ford, sun2020scalability}. In fact, though a single lidar mounted on the top of the vehicle can be cost sufficient, there are many scenarios where the use of multiple lidars can be advantageous.
For example, in some complex urban environments \cite{jiao2020robust, jeong2019complex}, the combined FOV from multiple lidars provides a more comprehensive observation of the surrounding, which allows the vehicle to navigate through the environment and avoid obstacles more flexibly.
In a challenging subterranean scenario \cite{palieri2020locus}, where the robot has to travel through some narrow and irregular terrains that can cause the lidar's FOV to be blocked, a multi-lidar based localization technique was shown to be the solution to overcome such challenge.

\begin{figure}[h]
		\centering
		\includegraphics[width=\linewidth]{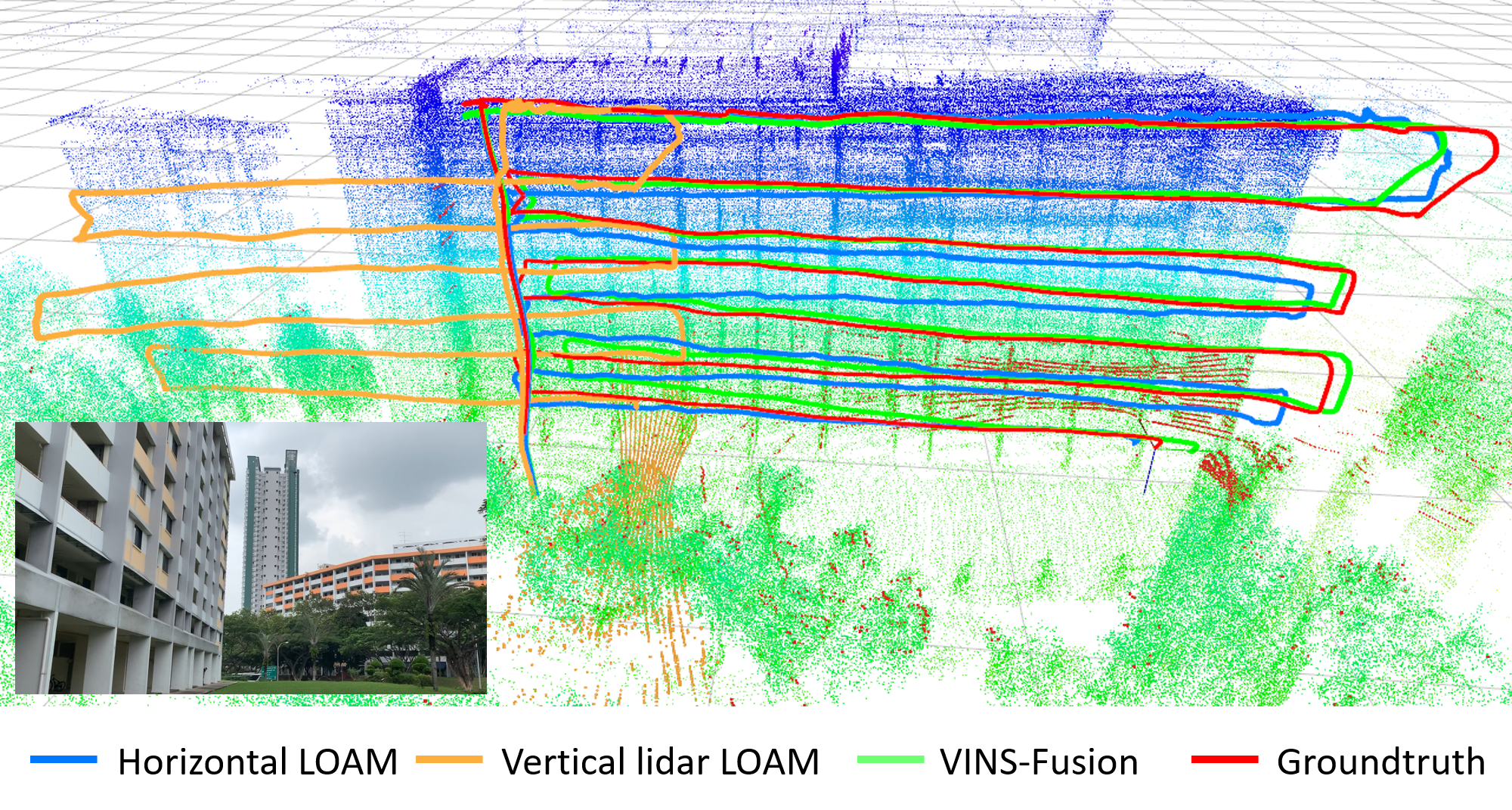}
		\caption{Illustration of the localization drift by using single lidar in a challenging environment.}
		\label{fig: aerial loam challenge}
\end{figure}

In the aforementioned works, we note that lidar was used mostly on ground robots operating in 2D space, hence it is usually assumed that the ground plane can be observed most of the time. This is no longer the case with aerial robots, which is the targeted application of this work. Fig. \ref{fig: aerial loam challenge} gives an example of such a scenario, in which, there are two lidars mounted on an Unmanned Aerial Vehicle (UAV), and the only nearby objects that can be observed are the building facade and the ground. Hence with a single horizontal lidar, a commonly known Lidar Odometry and Mapping (LOAM) method\footnote{\url{https://github.com/HKUST-Aerial-Robotics/A-LOAM}} gives the result that expectedly drifts in the vertical direction (blue path in Fig. \ref{fig: aerial loam challenge}), as it cannot observe the ground. Similarly, when using the vertical lidar, the result exhibits significant drift in the lateral direction (orange path Fig. \ref{fig: aerial loam challenge}), as it lacks observations to counter drift in that direction. In this work our main goal is to show that when these lidars are combined in a tighly coupled manner, the challenge can be effectively overcome.

\section{Related Works}

Many 3D lidar-based localization methods have been proposed in the last decade. Most notable is the work by Zhang et al \cite{zhang2014loam, zhang2018laser}. In \cite{zhang2014loam}, the authors proposed a standard framework for lidar-based odometry and mapping. Indeed, their technique to calculate smoothness and determine plane and edge features from lidar scans remains a popular approach that is still widely adopted to this day. In \cite{zhang2018laser}, the authors proposed a loosely coupled lidar-visual-inertial localization and mapping system, which was also demonstrated to work on a UAV. However, several techniques used in this work seem to be no longer efficient compared to current state-of-the-art. For example, the edge and plane features have separate factors with high non-linearity, while in recent works, it has been shown that one edge factor can be substituted by two plane factors \cite{jiao2020robust, shan2020liosam} with relatively simpler formulation. In addition, the voxel-based map management scheme appears to be quite complex. In contrast, inspired by contemporary VIO frameworks, in this work we propose a more efficient framework based on so-called \textit{key frame}, which simplifies the query as well as update on the map.

Besides, to the best of our knowledge, few works have satisfactorily investigated the tight coupling of IMU preintegration factors with lidar factors in the literature, despite its extensive applications in visual-inertial systems \cite{forster2016manifold, qin2017vins, nguyen2021viral, nguyen2020tightlyicra, nguyen2021range}.
In \cite{geneva2018lips}, a method based on this idea was proposed, however the lidar processing part assumes the presence of distinct planes that can be extracted via RANSAC in the environments.
In \cite{ye2019tightly}, a tightly-coupled framework called LIO-Mapping was proposed. Inspired by the VIO frameworks, both of the aforementioned works use a sliding window, and employ IMU preintegration and lidar features to construct cost factors that couple the measurements with the robot poses in the sliding window. 
Later, Shan et al. released the LIO-SAM package \cite{ye2019tightly} which optimizes a pose graph leveraging IMU preintegration and pose priors as the cost factors.
Unfortunately, LIO-Mapping performs very poorly when forced to run in real-time, which is also reported in other works \cite{nguyen2020liro, shan2020liosam, palieri2020locus}. We believe this is due to several inefficient implementation designs. On the other hand, LIO-SAM requires 9DoF IMU, and should be better categorized as a loosely coupled method, as it uses pose priors that are obtained from a LOAM process with a separate LM-based optimization process proposed earlier\cite{shan2018lego}, instead of jointly optimizing the lidar feature and IMU preintegration factors together.
\highlight{Besides, we also note that the aforementioned works do not generalize to multi-lidar case, which poses a new challenge as it requires us to design a stable synchronization scheme, as well as a time-efficient frontend to ensure real-time performance.} As will be shown later, despite integrating multiple lidars, and employing tightly coupled scheme, our method can still operate in real-time and provide accurate estimate, thanks to the use of multi-thread implementation and low programming overheads.

The main contribution of this work can be listed as follows:
\begin{itemize}
    \item We propose a general scheme to combine multiple lidars with complementary FOV for feature-based Lidar-Inertia Odometry and Mapping (LIOM) application.
    \item We propose a tightly-coupled, key frame-based, multi-threaded multi-input LIOM framework to achieve robust localization estimate.
    \item We demonstrate the advantages of the method over existing methods via extensive experiments on an advanced UAV platform.
\end{itemize}

The remainder of the paper is organized as follows: in Sec. \ref{sec: preliminary}, we lay out the basic definitions, notations and formulations of the problem; Sec. \ref{sec: overview}
then presents an overview of the software structure and the flow of information. Sec. \ref{sec: lidar proccessing} and Sec. \ref{sec: imu processing} discuss in details how lidar data and IMU are processed. Next we describe the construction of the local map and the \textit{feature-to-map matching} (FMM) process that produces the FMM coefficients for constructing the lidar factors. Sec. \ref{sec: map management} explains the key frame management procedures. We demonstrate the capability of our method via several experiments on UAV datasets in Sec. \ref{sec: experiment}. Finally, Sec. \ref{sec: conclusion} concludes our work.

\section{Preliminaries} \label{sec: preliminary}

\subsection{Nomenclature}

In this paper we use $(\cdot)^\top$ to denote the transpose of an algebraic vector or matrix under $(\cdot)$.
For a vector $ \vbf{x} \in \R^m $, $\norm{\vbf{x}}$ stands for its Euclidean norm, and $\norm{\vbf{x}}_{\vbf{G}}^2$ is  short-hand for 
$\norm{\vbf{x}}_{\vbf{G}}^2 = \vbf{x}^\top \vbf{G} \vbf{x}$.
For two vectors $\vel_1$, $\vel_2$, $\vel_1\times\vel_2$ denotes their cross product. In later parts, we denote $\rot \in \SO$ as the rotation matrix, and $\tf \in \SE$ as the transformation matrix.
We denote $\Q$ as the set of unit quaternions, with $\mathbb{q}$ as its identity element. Given $\quat \in \Q$, $\qtoR(\quat)$ denotes its corresponding rotation matrix, and $\vbf{vec}(\quat)$ returns its vector part. Also, in the later parts, we refer to the mapping $\calE \colon \R^3 \mapsto \Q$ and its inverse $\calE^{-1} \colon \Q \to \R^3$ to convert a \textit{rotation vector}, i.e. the {angle-axis representation} of rotation, to its quaternion form, and vice versa. The formulae for these operations can be found in \cite{nguyen2021viral}.

When needed for clarity, we attach a left superscript to an object to denote its coordinate frame. For example ${}^\fr{A}\mathbf{v}$ implies the coordinate of the vector $\mathbf{v}$ is in reference to the frame $\fA$, and ${}^{\fB}\F$ indicates that the coordinate of the points in the pointcloud $\F$ is \wrt to the frame $\fB$.
A rotation matrix and transformation matrix between two coordinate frames are denoted with the frames attached as the left-hand-side superscript and subscript, e.g., ${}^\fA_\fB\rot$ and ${^\fA_\fB\tf}$ are called the rotation and transform matrices \text{from frame $\fA$ to $\fB$}, respectively.
When the coordinate frames are the body frame at different times, depending on the context, we may omit and rearrange the superscript and subscripts to keep the notation concise, e.g., ${}^k\rot_{k+1} \triangleq {}^{\fB_k}_{\fB_{k+1}}\rot$, or ${}^w_m\tf \triangleq {}^{\fB_w}_{\fB_{m}}\tf$.

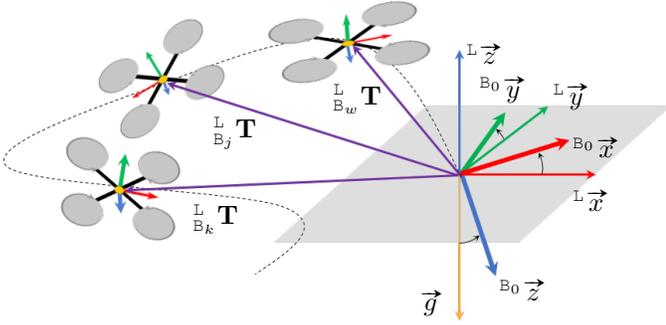
\begin{figure}[h]
    \centering
    \begin{overpic}[width=\linewidth,
                        ]{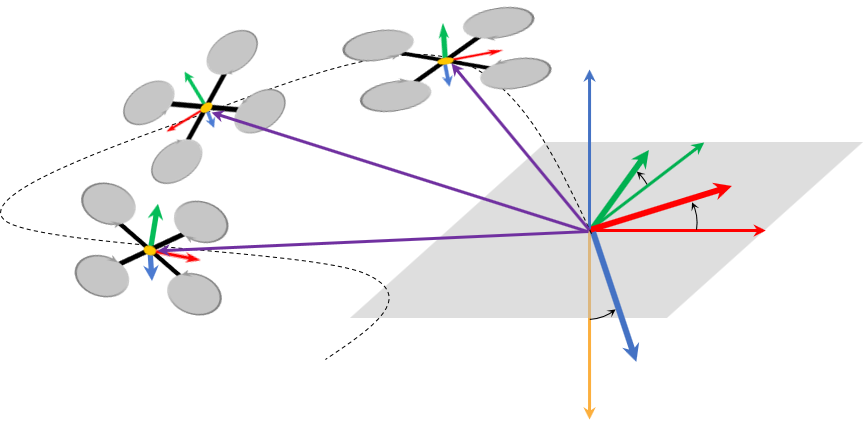}
                        \put(70.00, 35.00){ ${}^{\fB_0}\vv{y}$}
                        \put(81.00, 34.00){ ${}^{\fL}\vv{y}$}
                        \put(84.00, 18.00){ ${}^\fL\vv{x}$}
                        \put(84.00, 26.00){ ${}^{\fB_0}\vv{x}$}
                        \put(68.00, 40.00){ ${}^{\fL}\vv{z}$}
                        \put(73.00, 04.50){ ${}^{\fB_0}\vv{z}$}
                        
                        \put(61.00, 03.00){ $\vv{g}$}
                        
                        \put(48.00, 34.00){ ${}_{\fB_w}^{\fL}\tf$}
                        \put(30.00, 29.00){ ${}_{\fB_j}^{\fL}\tf$}
                        \put(27.00, 16.00){ ${}_{\fB_k}^{\fL}\tf$}
	\end{overpic}
	\caption{The local frame $\fL$ defined via the initial body frame $\fB_0$, and the poses of the robot at later time instances.}
	\label{fig: coordinates}
\end{figure}

Throughout this paper we define a local frame $\fL$ whose origin coincides with the position of the body frame at the initial time, denoted as $\fB_0$.
In addition, the ${}^{\fL}\vv{z}$ axis of $\fL$ points towards the opposite direction of the gravity vector $\vv{g}$ (see Fig. \ref{fig: coordinates}), its ${}^{\fL}\vv{x}$ axis points to the same direction of the projection of ${}^{\fB_0}\vv{x}$ on a plane perpendicular with ${}^\fL\vv{z}$ axis, and the ${}^\fL\vv{y}$ axis can be determined by the right-hand rule. Indeed, these calculations can be done by simply using the initial acceleration reading from the IMU, which contains only the coordinate of the vector $\vv{g}$ in $\fB_0$, to determine the pitch and roll angles of $\fB_0$ with respect to $\fL$, assuming that the error due to acceleration bias is negligible.


\subsection{State estimates}

In reference to Fig. \ref{fig: coordinates}, we define the robot states to be estimated at time $t_k$ as:
\begin{align}
    \X_k &= \Big( \quat_k,\ \pos_k,\ \vel_k,\ \bias^{\o}_{k},\ \bias^{a}_{k}\Big), \label{eq: state vector}
\end{align}
where $\quat_k \in \Q$, $\pos_k \in \R^3$, $\vel_k \in \R^3$ are respectively the orientation quaternion, position and velocity of the robot \wrt the local frame $\fL$ at time $t_k$; $\bias^{a}_{k},\ \bias^{\o}_{k} \in \R^3$ are respectively the IMU accelerometer and gyroscope biases.
Hence, we denote the state estimate at each time step $k$, and the sliding window of size $M$ as follows:
\begin{alignat}{2}
    &\Xhat_k &&= \Big( \hat{\quat}_k,\ \hat{\pos}_k,\ \hat{\vel}_k,\ \hat{\bias}^{\o}_{k},\ \hat{\bias}^{a}_{k}\Big), \label{eq: X hat k}\\
    &\Xhat &&= \l(\Xhat_{w},\ \Xhat_{w+1}, \dots,\ \Xhat_{k}\r),\ w \triangleq k - M + 1. \label{eq: X hat k to k+M}
\end{alignat}
Note that in this work we assume the extrinsic parameters have been manually calibrated, and each lidar's pointcloud has been transformed to the body frame before being processed further.
Moreover, depending on the context we may also refer to $\hat{\rot}_k \triangleq \qtoR(\hat{\quat}_k)$ as the rotation matrix estimate, and $(\hat{\rot}_k, \hat{\pos}_k)$ or $(\hat{\quat}_k, \hat{\pos}_k)$ as the pose estimate.

\section{General Framework} \label{sec: overview}

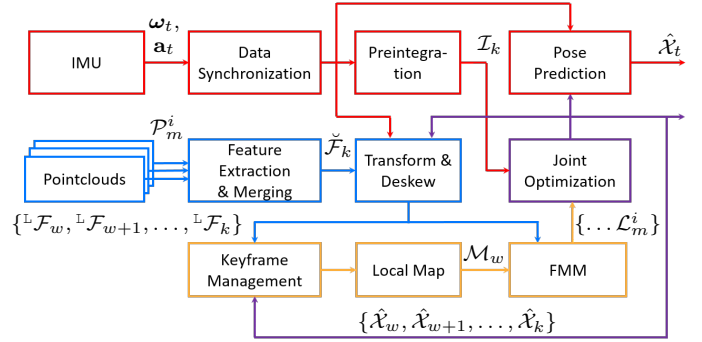
\begin{figure}[h]
    \centering
    \begin{overpic}[width=\linewidth,
                        ]{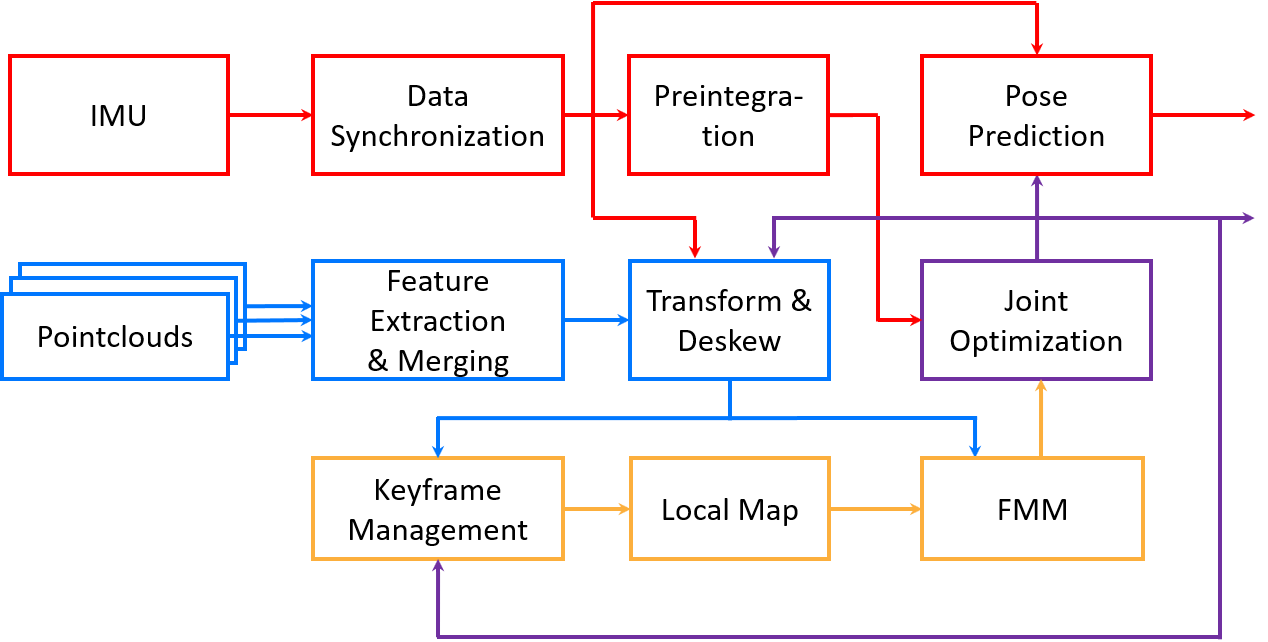}
                        \put(95.0,  43.0){\footnotesize  $\Xhat_{t}$}
                        \put(16.6,  45.0){\footnotesize  $\begin{array}{c}\angvel_t,\\\accel_t\end{array}$}
                        \put(18.8,  31.0){\footnotesize  $\P^i_m$}
                        \put(45.0,  28.0){\footnotesize  $\Fbrv_k$}
                        \put(68.0,  43.5){\footnotesize  $\I_k$}
                        \put(82.5,  16.5){\footnotesize  $\{\dots \L^i_m\}$}
                        \put(-2.0,  16.5){\footnotesize  $\{{}^\fL\F_w, {}^\fL\F_{w+1},\dots,{}^\fL\F_k\}$}
                        \put(65.80, 12.0){\footnotesize  $\M_w$}
                        \put(50.00, 1.75){\footnotesize  $\{\Xhat_{w}, \Xhat_{w+1}, \dots, \Xhat_{k} \}$}
	\end{overpic}
	\caption{Main processes and available quantities in the system by the time $t$. Note that lines of different colors are not connected.}
	\label{fig: overview}
\end{figure}

Fig. \ref{fig: overview} presents a snapshot of our MILIOM system at some IMU sample time $t$ past $t_k$, where $t_k$ is the start time of the \textit{skewed combined feature cloud} (SCFC) $\Fbrv_k$ (more details are given in Sec. \ref{sec: lidar proccessing}). Indeed, there are two types of features that can be extracted from the lidar scans, namely \textit{edge} feature and \textit{plane} feature. Thus when we refer to a feature cloud $\F$, we do mean the compound $\F \triangleq (\F^e,\ \F^p)$, where $\F^e$ is the set of edge features, and $\F^p$ is the set of plane features. Whenever the context requires, we can make the notation more explicit.

The inputs of the system are the IMU measurements, i.e. acceleration $\accel_t$ and angular velocity $\angvel_t$, and lidar raw pointclouds $\P_m^i$. The outputs are the optimized estimates of the sliding window state estimates, denoted as $\{\Xhat_{w}, \Xhat_{w+1}, \dots, \Xhat_{k}\}$, and the IMU-predicted state estimate up to time $t$, denoted as $\Xhat_t$.

In the above framework, the joint optimization process consumes the most computation time, while most other processes are meant to extract the key information used for constructing the factors that will be optimized in the following cost function:
\begin{align}
    f(\hat{\X}) &\triangleq
    \Bigg\{\mkern9mu \sum_{m = w+1}^{k} \norm{\resi_\I(\hat{\X}_{m-1}, \hat{\X}_{m}, \I_m)}^2_{\vbf{P}_{\I_m}^{-1}} \nonumber\\
    &\qquad+ \sum_{m = w}^{k} \sum_{i = 1}^{\abs{\C_m}}\rho\l(\norm{\resi_\L(\hat{\X}_{m}, \L_m^i)}^2_{\vbf{P}_{\L_m^i}^{-1}}\r)\Bigg\}, \label{eq: cost function}
\end{align}
where $\resi_\I(\cdot)$ and $\resi_\L(\cdot)$ are the IMU and lidar residuals, $\rho(\cdot)$ is the Huber loss function to suppress outliers, $\vbf{P}_{\I_m}$ and $\vbf{P}_{\L_m^i}$ are the covariance matrix of their measurements, $\I_m$ and $\L_m$ are respectively IMU preintegration and lidar FMM coefficients, $\C_m$ denotes the set of FMM features derived from the lidar feature cloud at time $t_m$. Note that each IMU factor in \eqref{eq: cost function} is coupled with two consecutive states, while each lidar factor is coupled with only one.
In this paper, the IMU factors are calculated via an integration scheme as described in our previous work \cite{nguyen2021viral}. Hence we only recall some important steps due to page constraints.
In the next sections we will elaborate on the processing of lidar and IMU data for use in the optimization of \eqref{eq: cost function}.

\section{Lidar processing} \label{sec: lidar proccessing}

\begin{figure}[h]
    \centering
    \begin{overpic}[width=\linewidth,
                    ]{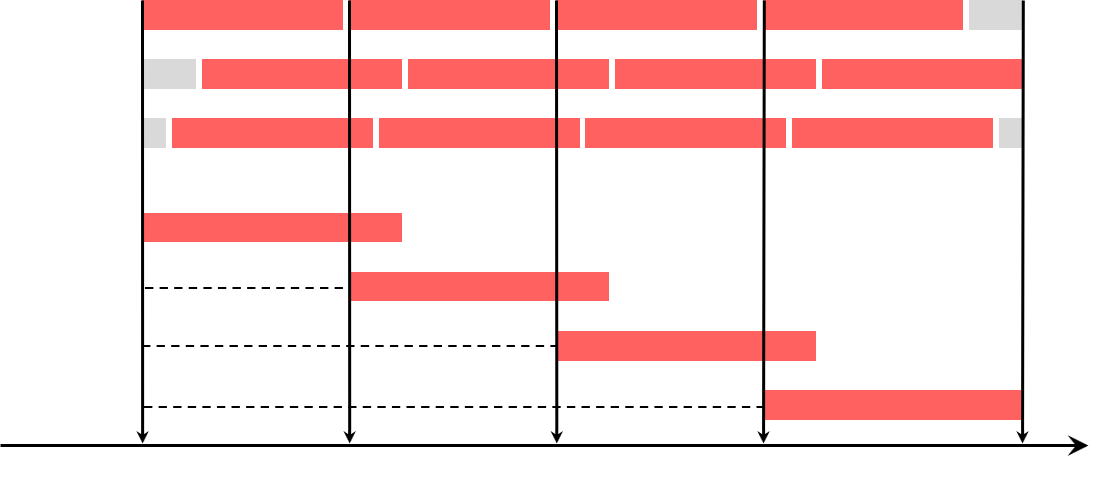}
                    \put(03.00,  42.00){\footnotesize $\P^1_m$}
                    \put(03.00,  36.50){\footnotesize $\P^2_m$}
                    \put(03.00,  31.00){\footnotesize $\P^n_m$}
                    
                    \put(02.00,  22.50){\footnotesize $\Fbrv_w$}
                    \put(02.00,  17.00){\footnotesize $\Fbrv_{w+1}$}
                    \put(02.00,  11.50){\footnotesize $\dots$}
                    \put(02.00,  06.50){\footnotesize $\Fbrv_{k}$}
                    
                    \put(10.50,  00.00){\footnotesize $t_w$}
                    \put(29.10,  00.00){\footnotesize $\dots$}
                    \put(49.00,  00.00){\footnotesize $t_{k-1}$}
                    \put(67.00,  00.00){\footnotesize $t_k$}
                    
                    \put(91.00,  00.00){\footnotesize $t_k'$}
	\end{overpic}
	\caption{Illustration of the feature extraction and synchronization between the lidar scans. Assuming that there are $n$ lidars, $\P^i_m$ refers to the pointclouds of lidar $i \in \{1, 2, \dots, n\}$. Note that $\Fbrv_{k}$ is the last SCFC that comes out from the feature extraction and synchronization block with the start time $t_k$ and the end time $t_k'$.}
	\label{fig: lidar lidar sync}
\end{figure}
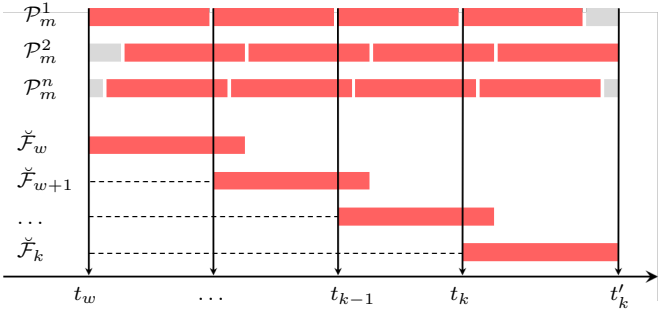

Fig. \ref{fig: lidar lidar sync} presents an illustration of the synchronization between the lidars.
Specifically, for a lidar $i$, we can obtain a sequence of pointclouds $\P^i_m$, each corresponds to a 3D scan of the environment over a fixed period (which is around 0.1s for the Ouster\footnote{\url{https://ouster.com/products/os1-lidar-sensor/}} sensor used in this work).
We assume that lidar 1 is the \textit{primary} unit, whose start times are used to determine the time instances of the state estimates on the sliding window, and others are referred to as \textit{secondary} lidars.
Each pointcloud input is put through a feature extraction process using the smoothness criteria as in \cite{zhang2014loam} to obtain the corresponding feature cloud. We then merge all feature clouds whose start times fall in the period $[t_{k-1}, t_{k})$ to obtained the SCFC $\Fbrv_k$.

Note that if the scan period of a lidar is 0.1s, then a SCFC can stretch over a maximum of 0.2s period. This is a significant time-span where the robot motion can affect the pointcloud in the scan, however since all of the points in the pointclouds are timestamped, we can use the IMU-based state propagation to compensate for the robot motion, which we refer to as the "deskew" process.
Specifically, for a feature point ${}^{\fB_{t_s}}\vbf{f}$ in the SCFC $\Fbrv_k$ stretching over the time span $[t_k,\ t_k')$, its coordinates \wrt the robot's pose at time $t_{k}$ can be calculated by the interpolation:
\begin{equation}
    {}^{\fB_{t_s}}\vbf{f} \triangleq \qtoR\l(\mathrm{slerp}\l(\mathbb{q}, {}^{t_k}_{t_k'}\breve{\quat}, s\r)\r){}^{\fB_{t_s}}\vbf{f} + s {}^{t_k}_{t_k'}\breve{\pos},
\end{equation}
where $s = \frac{t_s}{t_k' - t_k}$ and $\mathrm{slerp}(\quat_1, \quat_2, s)$ is the \textit{spherical linear interpolation} operation on quaternions \cite{dam1998quaternions}, ${}^{t_k}_{t_k'}\breve{\quat}$ and ${}^{t_k}_{t_k'}\breve{\pos}$ are the orientation and position of $\fB_{t_k'}$ \wrt $\fB_{t_k}$. More information on the IMU propagation is given in Sec. \ref{sec: imu propagation}.

After deskew, a SCFC $\Fbrv_m$ becomes ${}^\fB\F_m$, the so-called \textit{deskewed combined feature cloud}, or simply CFC for short. We further use the state estimates $\{\Xhat_{w}, \Xhat_{w+1}, \dots, \Xhat_{k-1}\}$ and the IMU-propagated states $\breve{\quat}_k$, $\breve{\pos}_k$ (see Fig. \ref{sec: imu propagation}) to transform $\{{}^\fB\F_w, {}^\fB\F_{w+1}, \dots, {}^\fB\F_{k}\}$ to $\{{}^\fL\F_w, {}^\fL\F_{w+1}, \dots, {}^\fL\F_{k}\}$, which are employed in the construction of the local map and the FMM process in Sec. \ref{sec: local map and FMM}.

\begin{remark}
    Our synchronization scheme differs from LOCUS \cite{palieri2020locus} in that the merging of the pointclouds is made on full-size poinclouds in \cite{palieri2020locus}, while ours is made on feature pointclouds. The reason for choosing this merging scheme is because the feature extraction part is done on \highlight{distinct \textit{rings}}, using the information on the horizontal and vertical angular resolution of the lidar sensors. Hence if the merging happens before the feature extraction, the \textit{ring by ring} structure will be lost and one cannot trace the smoothness of the points on the ring to extract feature. On the other hand, LOCUS uses General Iterative Closest Point (GICP) for scan matching, which does not concern features, however it requires good IMU and/or wheel-inertial odometry (WIO) information for initial guess of the relative transform between the two pointclouds.
\end{remark}


\section{IMU Processing} \label{sec: imu processing}

\begin{figure}[h]
    \centering
    \begin{overpic}[width=\linewidth,
                    ]{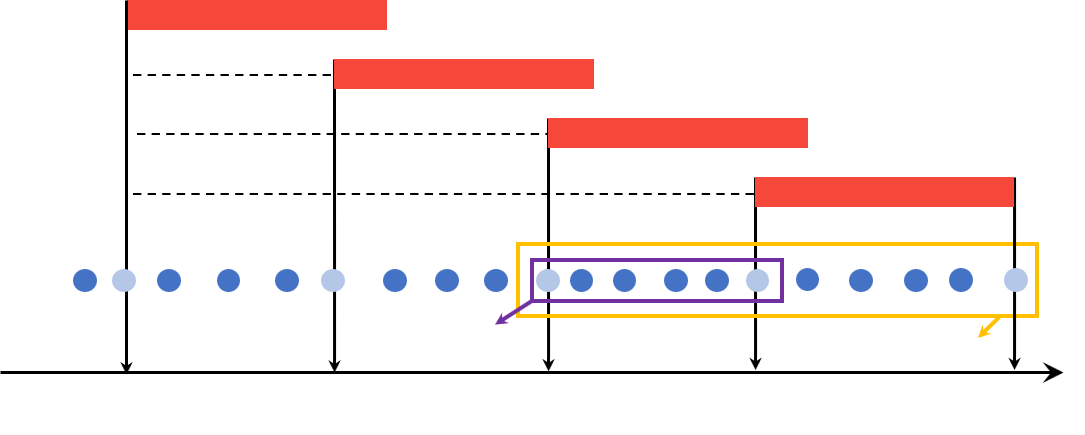}
                    
                    \put(00.00,  38.00){\footnotesize ${}^\fL\F_w$}
                    \put(00.00,  33.00){\footnotesize ${}^\fL\F_{w+1}$}
                    \put(00.00,  27.00){\footnotesize ${}^\fL\F_{k-1}$}
                    \put(00.00,  21.50){\footnotesize $\Fbrv_{k}$}
                    
                    \put(00.00,  13.0){\footnotesize IMU}
                    
                    \put(31.40,  07.00){\footnotesize Preintegration}
                    \put(75.00,  07.00){\footnotesize Propagation}
                    
                    \put(09.00,  01.50){\footnotesize $t_w$}
                    \put(30.00,  01.50){\footnotesize $\dots$}
                    \put(48.75,  01.50){\footnotesize $t_{k-1}$}
                    \put(69.00,  01.50){\footnotesize $t_k$}
                    
                    \put(93.00,  01.00){\footnotesize $t_k'$}
	\end{overpic}
	\caption{Synchronization between the CFC and the IMU. The light blue circles represent the interpolated samples at the start and end times of the CFCs.}
	\label{fig: lidar imu sync}
\end{figure}
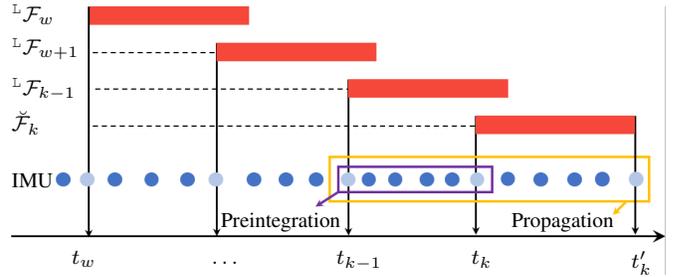

In Fig. \ref{fig: lidar imu sync}, we illustrate the process of synchronizing the IMU data with the lidar pointcloud. Recall that $t_k'$ is the end time of the last SCFC, and that during the period $[t_{k-1},\ t_k')$, we obtain the following IMU samples $\{(\angvel_{\tau_m}, \accel_{\tau_m}), (\angvel_{\tau_{m+1}}, \accel_{\tau_{m+1}}), \dots, (\angvel_{\tau_{N'}}, \accel_{\tau_{N'}})\}$. These samples will be used for two operations below.

\subsection{Preintegration} \label{sec: preintegration}
First, we use the samples in the sub-interval $[t_{k-1},\ t_k)$ to calculate the so-called \textit{preintegration observations} using the zero-order-hold (ZOH) technique:
\begin{subequations} \label{eq: pIMU zoh}
    \begin{align}[left = \empheqlbrace]
      &\breve{\pig}_{\tau_{m+1}} = \breve{\pig}_{\tau_{m}}
      \circ
      \calE\l(\D\tau_{m}\bar{\angvel}_{\tau_{m}}\r), \label{eq: pIMU gamma zoh rule}
      \\
      &\breve{\pib}_{\tau_{m+1}} = \breve{\pib}_{\tau_m} + \D\tau_{m}\qtoR(\breve{\pig}_{\tau_m})\bar{\accel}_{\tau_m}, \label{eq: pIMU beta zoh rule}
      \\
      &\breve{\pia}_{\tau_{m+1}} = \breve{\pia}_{\tau_m} + \D \tau_{m}\breve{\pib}_{\tau_m} + \frac{\D\tau_{m}^2}{2}\qtoR(\breve{\pig}_{\tau_m}) \bar{\accel}_{\tau_m}, \label{eq: pIMU alpha zoh rule}
      \\
      & \breve{\pig}_{\tau_{0}} \triangleq \q,\ \breve{\pib}_{\tau_{0}} \triangleq \vbf{0},\ \breve{\pia}_{\tau_{0}} \triangleq \vbf{0}, \label{eq: pimu preint init}
      \\
      & \bar{\angvel}_{\tau_{m}} \triangleq ({\angvel}_{\tau_m} - \hat{\bias}^\o_{k-1}),\ \bar{\accel}_{\tau_m} \triangleq ({\accel}_{\tau_m} - \hat{\bias}^a_{k-1}), \label{eq: imu measurement no bias}
      \\
      &\D \tau_m \triangleq \tau_{m+1} - \tau_{m},\ m \in \{0, 1, \dots, N\},
      \\
      &\tau_{0} = t_{k-1},\ \tau_{N+1} = t_k,
    \end{align}
\end{subequations}
where $\circ$ is the quaternion product, and $\calE(\cdot)$ is the mapping from rotation vector to quaternion described in Sec. \ref{sec: preliminary}.

At the end of the preintagration process \eqref{eq: pIMU zoh}, we can obtain the preintegration observations $\I_k \triangleq (\breve{\pia}_{t_k}, \breve{\pib}_{t_k}, \breve{\pig}_{t_k})$.
The use of this observation in an optimization framework has been described in depth in our previous work \cite{nguyen2021viral}, and we refer to that for more details.

\subsection{State Propagation} \label{sec: imu propagation}
Given the IMU samples over $[t_{k-1},\ t_k')$ and the state estimate $\Xhat_{k-1}$, we propagate the state further in time and obtain the IMU-predicted poses at time $t_k$ and $t_k'$, denoted as $(\breve{\pos}_k, \breve{\quat}_k)$ and $(\breve{\pos}_{t_k'}, \breve{\quat}_{t_k'})$, for the deskew of the SCFC. The propagation is done using the ZOH technique as follows:
\begin{subequations} \label{eq: propagate}
    \begin{align}[left = \empheqlbrace]
      &\breve{\quat}_{\tau_{m+1}} = \breve{\quat}_{\tau_{m}}
      \circ
      \calE\l(\D\tau_m\bar{\angvel}_{\tau_m}\r), \label{eq: propagate quat}
      \\
      &\breve{\vel}_{\tau_{m+1}} = \breve{\vel}_{\tau_{m}} + \D\tau_m\l[\qtoR(\breve{\quat}_{\tau_m})\bar{\accel}_{\tau_m} - \grav\r], \label{eq: propagate vel}
      \\
      &\breve{\pos}_{\tau_{m+1}} = \breve{\pos}_{\tau_{m}} + \D\tau_m\breve{\vel}_{\tau_m} \dots \nonumber
      \\ 
      &\qquad\qquad\quad\mkern12mu+ \frac{(\D\tau_m)^2}{2}\l[\qtoR(\breve{\quat}_{\tau_m})\bar{\accel}_{\tau_m} - \grav\r], \label{eq: propagate pos}
      \\
      &\breve{\quat}_{\tau_0} = \hat{\quat}_{k-1},\ \breve{\vel}_{\tau_0} = \hat{\vel}_{k-1},\ \breve{\pos}_{\tau_0} = \breve{\pos}_{k-1}, \label{eq: propagate init}
      \\
      &\D \tau_m \triangleq \tau_{m+1} - \tau_{m},\ m \in \{0, 1, \dots, N, \dots, N'\},
      \\
      &\tau_{0} = t_{k-1},\ \tau_{N+1} = t_k,\  \tau_{N'+1} = t,
    \end{align}
\end{subequations}
where we reuse the definition of $\bar{\angvel}_{\tau_m}$ and $\bar{\accel}_{\tau_m}$ in \eqref{eq: imu measurement no bias}.

Hence we obtain the relative transform from time $t_k$ to $t_k'$ by ${}^{t_k}_{t_{k}'}\breve{\quat} \triangleq \breve{\quat}_{t_k}^{-1}\breve{\quat}_{t_k'}$ and ${}^{t_k}_{t_{k}'}\breve{\pos} \triangleq -\qtoR(\breve{\quat}_{t_k}^{-1})(\breve{\quat}_{t_k'} -\breve{\quat}_{t_k})$. This transform is used for the deskew of the SCFC in Sec. \ref{sec: lidar proccessing}.

\begin{remark}
It might be useful to note the difference between IMU propagation and IMU preintegration. On the physical sense, IMU propogation uses IMU measurement to project the system's state from time $t_k$ to time $t$, while preintegration is the direct \textit{integration} of IMU measurements to obtain some pseudo-observation \cite{lupton2011visual}. In a technical sense, preintegration does not involve cancelling the gravity from the acceleration measurement before integration (when comparing \eqref{eq: pIMU beta zoh rule}, \eqref{eq: pIMU alpha zoh rule} with \eqref{eq: propagate vel}, \eqref{eq: propagate pos}), nor does it require the initial state $(\hat{\quat}_{k-1}, \hat{\vel}_{k-1}, \breve{\pos}_{k-1})$ (when comparing \eqref{eq: pimu preint init} with \eqref{eq: propagate init}).
\end{remark}

\section{Local Map and FMM Processes} \label{sec: local map and FMM}

The local map $\M_w$ is used as a prior map to calculate the FMM coefficients for the joint optimization process. In the beginning, when fewer than $M$ key frames have been stored to the memory, we directly merge the CFCs of the first $M-1$ steps obtained in Sec. \ref{sec: lidar proccessing} to obtain $\M_w$. On the other hand, when enough key frames have been created, we will use the latest IMU-predicted state estimate $\breve{\pos}_{k}$ to search for the $M$ nearest key frames, and merge the corresponding pointclouds to obtain $\M_w$.

Next, for each CFC ${}^\fL\F_m$, $m=w, w+1, \dots, k$, we calculate the FMM coefficients for each of its features by using Algorithm \ref{algo: lidar coefficients}.
In steps \ref{algo: plane feature start}-\ref{algo: plane feature end} of this algorithm, for each plane feature $\f$ we find a set $\mathcal{N}_\f$ of neighbouring plane features in the local map using KNN, then calculate the Hesse normal $\n$ of a plane, whose sum of squared distances to the points in $\mathcal{N}_\f$ is minimal (this optimization problem has closed-form solution). We then calculate the "fitness score" $s$ in step \ref{algo: plane finess score}. If the fitness score is above a threshold in, we can admit the tuple $({}^{m}\f^i, g\bar{\n}, g)$ to the set of FMM coefficients of CFC $\F_m$, denoted as $\C_m$.

\begin{figure}
    \centering
    \includegraphics[width=0.9\linewidth]{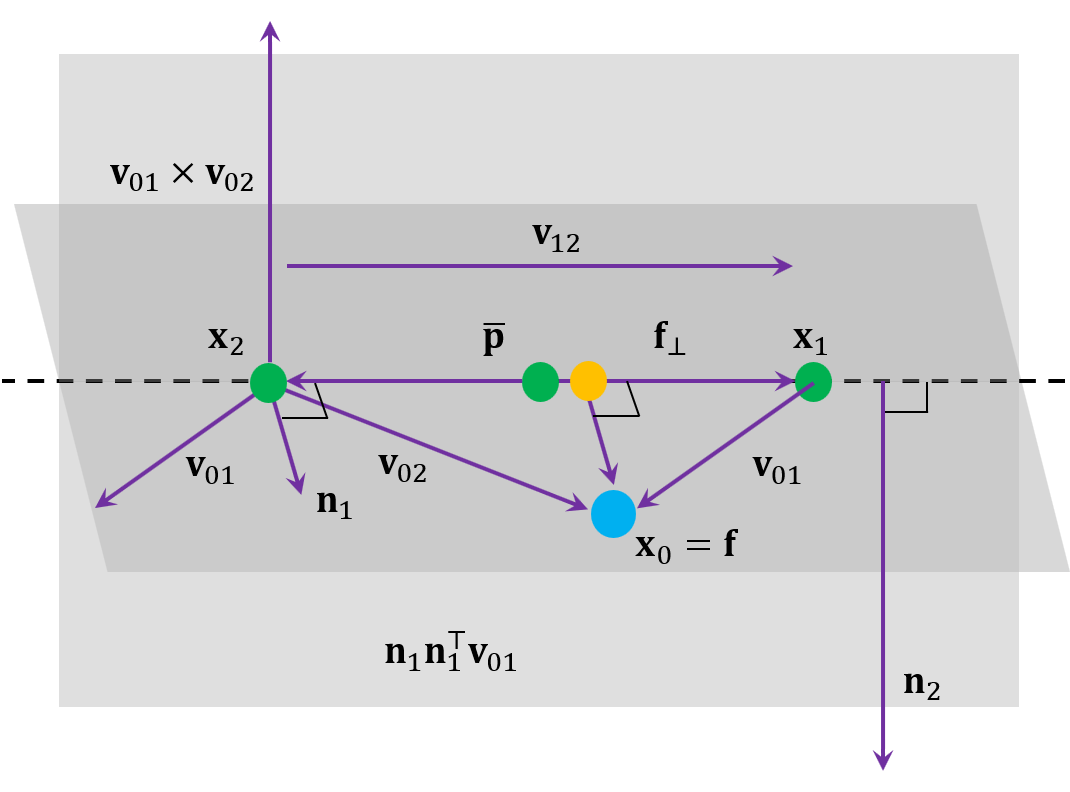}
    \caption{Illustrations of the vectors and the two planes computed in step \ref{algo: planes start} to step \ref{algo: planes end} of Algorithm \ref{algo: lidar coefficients}.}
    \label{fig: vectors}
\end{figure}

For steps \ref{algo: edge feature start}-\ref{algo: edge feature end}, we follow a similar procedure to calculate the coefficients of an edge feature. Our strategy here is to construct the two planes that intersect at the presumed edge line, as such, if $\f$ belongs to the edge line, it belongs to both of these planes and vice versa. To find the edge line, we find the fittest line going through the neighbor set $\mathcal{N}_\f$ and its centroid $\bar{\pos}$, which corresponds to steps \ref{algo: centroid start}-\ref{algo: centroid end}. From step \ref{algo: planes start} to step \ref{algo: planes end}, using some geometrical manipulations, we compute the Hesse normal vectors of the two planes, one that goes through this line and $\f$, and another that also goes through the edge line but is perpendicular to the first plane. Hence, we compute the fitness score to decide whether to add the tuples $({}^{m}\f^i, g\bar{\n}_1, g c_1)$ and $({}^{m}\f^i, g\bar{\n}_2, g c_2)$ to the set $\C_m$. Fig. \ref{fig: vectors} illustrates the vectors and planes computed over these steps.

It should be noted that the FMM process can be quite computationally expensive. However, since Algorithm \ref{algo: lidar coefficients} does not modify the local map, the FMM process can be split into multiple threads for each CFC. Thanks to this strategy, we can cut down the computation time to about 10m to 20ms compared to hundreds of ms when using a single thread.

\begin{algorithm}
  \SetAlgoLined
  \KwIn{$\F_{m} = (\F_{m}^p,\ \F_{m}^e)$, $\M_w = (\M_{w}^p,\ \M_{w}^e)$, $\hat{\tf}_m$.}
  \KwOut{$\C_m \triangleq \{\L^1_m, \L^2_m, \dots, \}$, $\L^i_m \triangleq ({}^{m}\f^i, \n^i, c^i)$.}
  \For{$\mathrm{each}$ $\f \in \F_m$}
  {
    Compute ${}^{m}\f^i$ from $\f^i$ and $\hat{\tf}_m$\;
    \uIf{ $\f \in \F_{m}^p$ }
    {\label{algo: plane feature start}
        Find $\mathcal{N}_{\f} = \text{KNN}(\f,\ \M_w^p)$\;
        Find $\bar{\n} = \argmin_{\n} \sum_{\vbf{x} \in \mathcal{N}_\f} ||\n^\top \vbf{x} + 1||^2$\;
        Compute $s = \l[1 - 0.9\frac{\abs{\bar{\n}^\top\f + 1}}{\norm{\bar{\n}}\norm{\f}}\r]$ and $g=\frac{s}{\norm{\bar{\n}}}$\;
        \label{algo: plane finess score}
        If $s > 0.1$, add $({}^{m}\f^i, g\bar{\n}, g)$ to $\C_m$\;
    \label{algo: plane feature end}
    }
    \uElseIf{$\f \in \F_{m}^e$}
    {\label{algo: edge feature start}
        Find the set $\mathcal{N}_\f = \text{KNN}(\f, \M_w^e)$, and its centroid $\bar{\pos} = \frac{1}{\abs{\mathcal{N}_\f}}\sum_{\vbf{x}\in\mathcal{N}_\f}\vbf{x}$\;
        \label{algo: centroid start}
        Compute: $\vbf{A} \triangleq \frac{1}{\abs{\mathcal{N}_\f}}\sum_{\vbf{x}\in\mathcal{N}_\f}(\vbf{x} - \bar{\pos})(\vbf{x} - \bar{\pos})^\top$\;
        Find the eigenvector $\vel_{\max}$ corresponding to the largest eigenvalue of $\vbf{A}$\;
        \label{algo: centroid end}
        Compute: $\vbf{x}_0 = \f$, $\vbf{x}_1 = \bar{\pos} + 0.1\vel_{\max}$, $\vbf{x}_2 = \bar{\pos} - 0.1\vel_{\max}$, $\vbf{v}_{01} = \vbf{x}_{0} - \vbf{x}_{1}$, $\vbf{v}_{02} = \vbf{x}_{0} - \vbf{x}_{2}$, $\vbf{v}_{12} = \vbf{x}_{1} - \vbf{x}_{2}$, \;
        \label{algo: planes start}
        Compute: $\n_1 = \vbf{v}_{12} \times (\vbf{v}_{10} \times \vbf{v}_{02})$,  $\n_1 \leftarrow \n_1/\norm{\n_1}$, $\n_2 = \vbf{v}_{12}\times\n_1$\;
        Compute: $\f_\bot = \f - (\n_1\n_1^\top)\vbf{v}_{01}$\;
        Compute: $c_1 = -\n_1^\top\f_\bot$ and $c_2 = -\n_2^\top\f_\bot$\;
        \label{algo: planes end}
        Compute: $s = \l[1 - \frac{0.9\norm{\vbf{x}_{01}\times\vbf{x}_{02}}}{\norm{\vbf{x}_{12}}}\r]$, $g = s/2$\;
        If $s > 0.1$, add $({}^{m}\f^i, g\bar{\n}_1, g c_1)$ and $({}^{m}\f^i, g\bar{\n}_2, g c_2)$ to $\C_m$\;
        \label{algo: edge feature end}
    }
  }
  \caption{Calculation of FMM coefficients}
  \label{algo: lidar coefficients}
\end{algorithm}

The result of the FMM is a list of coefficients used to create a lidar factor in \eqref{eq: cost function}. Specifically, for each tuple of FMM coefficients $\L^i_m = ({}^{m}\f^i, \n^i, c^i)$, we can calculate the corresponding residual as follows:
\begin{equation}
    \resi_\L(\hat{\X}_{m}, \L_m^i) = (\n^i)^\top\l[\qtoR(\hat{\quat}_m){}^{m}\f^i + \hat{\pos}_m\r] + c^i.
\end{equation}

\begin{remark}
Note that both ${}^\fL\F_m$ and $\M_w$ in Algorithm \ref{algo: lidar coefficients} are \wrt to the $\fL$ frame, thus we omit the superscript for more concise notation.
Also, either $\M_w$ or $\F_m$ or both can be down-sampled to reduce the computational load. In this paper we chose a 0.4m leaf size for the plane feature pointclouds and 0.2m for the edge feature pointclouds.
\end{remark}

\section{Key frame Management} \label{sec: map management}

The key frame in this case refers to marginalized pose estimates, so-called key poses, and their corresponding CFCs, so-called key pointclouds.
After each joint optimization step, we will consider admitting the middle pose estimate, i.e. $(\hat{\quat}_v, \hat{\pos}_v),\ v \triangleq k - M/2$ to the list of key poses. Specifically, given $\hat{\pos}_v$, we will find its KNN among the key poses and create the sets $\mathcal{N}_p = \{\pos_0, \pos_1, \dots, \pos_K\}$ and $\mathcal{N}_q = \{\quat_0, \quat_1, \dots, \quat_K\}$ that contain the positions and quaternions of these key poses, respectively.
Hence, we check the following:
\begin{itemize}
    \item $\norm{\hat{\pos}_v - \pos} > 1.0,\ \forall \pos \in \mathcal{N}_p$,
    \item $\norm{\calE^{-1}(\hat{\quat}_v^{-1}\quat)} > \pi/18,\ \forall \quat \in \mathcal{N}_q$,
\end{itemize}
where $\calE^{-1}$ is the inverse of $\calE$, which returns the rotation vector from a quaternion.
If either one of the two conditions above is true, then we will admit this pose estimate to the list of key poses and save its corresponding CFC $\F_v$ to the buffer for later use in the construction of local map.

\section{Experiment} \label{sec: experiment}

In this section we describe our implementation of the MILIOM method and seek to demonstrate its effectiveness through a series of experiments on UAV datasets.
The software system is developed on the ROS framework, employs the ceres-solver for the joint optimization process and other libraries such as PCL, Eigen, etc for their utilities. All experiments were done on a core-i7 computer with 6 cores. Video recording of some experiments can be viewed at \url{https://youtu.be/dHXYWC2KJyo}.

\subsection{NTU VIRAL Datasets}
\begin{figure}[t]
    \centering
    \includegraphics[width=\linewidth]{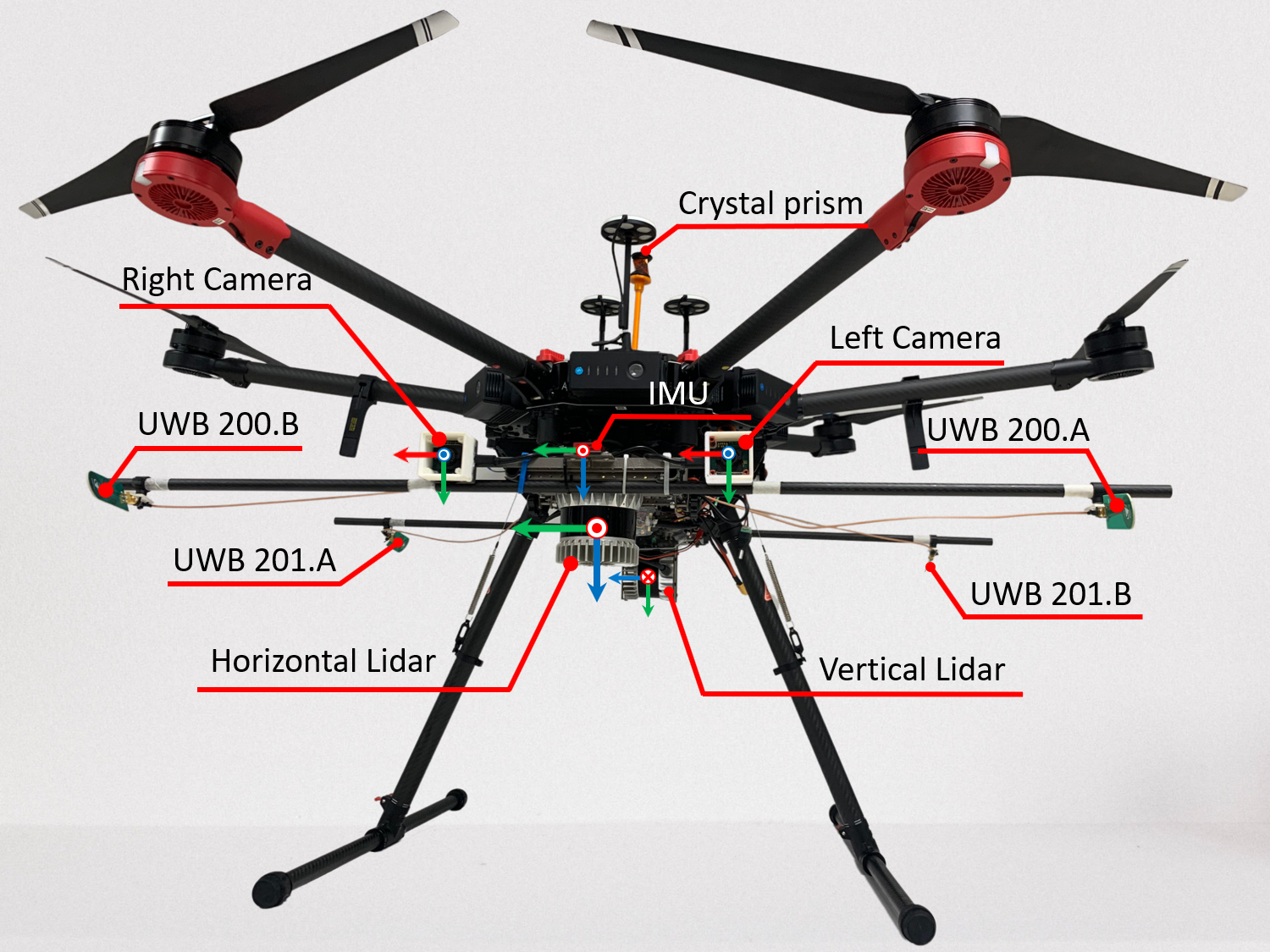}
    \caption{Hardware setup of the viral dataset, with one horizontal and one vertical lidar whose fields of view are complementary.}.
    \label{fig: hardware}
\end{figure}

We employ our recently published NTU VIRAL dataset \cite{nguyen2021ntuviral}\footnote{\url{https://ntu-aris.github.io/ntu\_viral\_dataset/}}, which to the best of our knowledge is the first UAV dataset that contains data from multiple lidars (besides multiple cameras, IMUs, Ultra-wideband (UWB) ranging sensors). The configuration of the two lidars can be seen in Fig. \ref{fig: hardware}. More specifically, one so-called horizontal lidar can scan the front, back, left, right, while the vertical lidar can scan the front, back, above and below sides of the UAV. \highlight{Both lidars have 16 channels with $32^o$ vertical FOV.}

We compare our algorithms with the two latest LIO based methods, which is LIO-Mapping \cite{ye2019tightly} (henceforth referred to as LIO-M) and LIO-SAM \cite{shan2020liosam}, \highlight{and the MLOAM\footnote{\url{https://github.com/gogojjh/M-LOAM}} \cite{jiao2020robust} method, which integrates multiple lidars but does not employ IMU measurements}. We note that both LIO-M and LIO-SAM are designed for single lidar configuration. More specifically, they assume the pointcloud input to be of regular shape and use the vertical and angular steps to extract the pointcloud features. In theory, we can modify their code to replace the internal feature extraction process with our \textit{feature extraction \& merging} part (Fig. \ref{fig: overview}) and inject the CFC into their backend. However this would involve too much modification into the original code structure. Therefore, we opt to configure our method as well as the others to work with a single horizontal lidar and report the result in Tab. \ref{tab: ATE}. We then configure our method to work with both lidars and append the result to Tab. \ref{tab: ATE}. \highlight{The details of these configurations will be provided on the NTU VIRAL dataset's website.}

\begin{table}[t]
\centering
\caption{\highlight{ATE of the lidar-based localization methods over the NTU VIRAL datasets. The best result is highlighted in \textbf{bold}, the second best is \ul{underlined}. All values are in m.}} \label{tab: ATE}
\renewcommand{\arraystretch}{1.1}
\begin{tabular}{c|ccc|cc}
\hline\hline
\mc{1}{c|}{\bf{Dataset}}
        & \mc{3}{c|}{\bf 1-Lidar}
        & \mc{2}{c}{\bf 2-Lidar}\\\cline{2-6}
        & \bf{LIO-M} & \bf{LIO-SAM} &\bf{Ours} &\bf{MLOAM} &\bf{Ours}\\\hline
eee\_01 &1.0542     & \ul{0.0915}   & 0.1042        &0.3558 & \bf{0.0666}   \\
eee\_02 &0.7234     & 0.0815        & \bf{0.0650}   &0.1945 & \ul{0.0656}   \\
eee\_02 &1.0314     & 0.1176        & \ul{0.0628}   &0.2996 & \bf{0.0518}   \\
nya\_01 &2.2436     & 0.0899        & \ul{0.0832}   &0.1555 & \bf{0.0565}   \\
nya\_02 &1.9664     & 0.1068        & \ul{0.0721}   &0.2334 & \bf{0.0668}   \\
nya\_03 &2.9934     & 0.3655        & \ul{0.0577}   &0.2859 & \bf{0.0423}   \\
sbs\_01 &1.6737     & 0.0966        & \ul{0.0764}   &0.1925 & \bf{0.0658}   \\
sbs\_02 &1.8056     & 0.0961        & \bf{0.0806}   &0.1778 & \ul{0.0816}   \\
sbs\_03 &2.0006     & 0.0960        & \bf{0.0884}   &0.1863 & \ul{0.0933}   \\\hline\hline
\end{tabular}
\end{table}

All parameters such as the standard deviation of IMU noises, window size, downsampling resolution, etc., are kept uniform. \highlight{All methods are required to run at the full lidar rate, which is 10 Hz}. However, for LIO-M, we reduce the rate of the lidar topic to one third of the actual rate (10 Hz) to allow the algorithm to run in real-time. Even with this we also have to limit the time for the optimization process when the buffered data becomes too large.
We run each algorithm on each dataset once and record the IMU-predicted estimates. The estimated and grountruth trajectories are then synchronized and aligned with each other. The resulting Absolute Trajectory Error (ATE) of each test and each method is computed and reported in Tab. \ref{tab: ATE}. These tasks are done by the matlab scripts accompanying the NTU VIRAL data suite \cite{nguyen2021ntuviral}.

	
	


\begin{figure}[t]
    \centering
    \includegraphics[width=\linewidth]{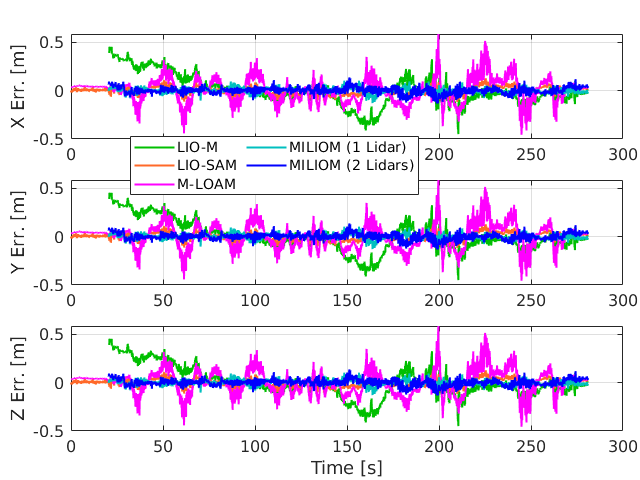}
    \caption{Estimation error by different methods on the eee\_02 dataset.}
    \label{fig: eee 02 err}
\end{figure}

\begin{figure}[h]
    \centering
    \includegraphics[width=\linewidth]{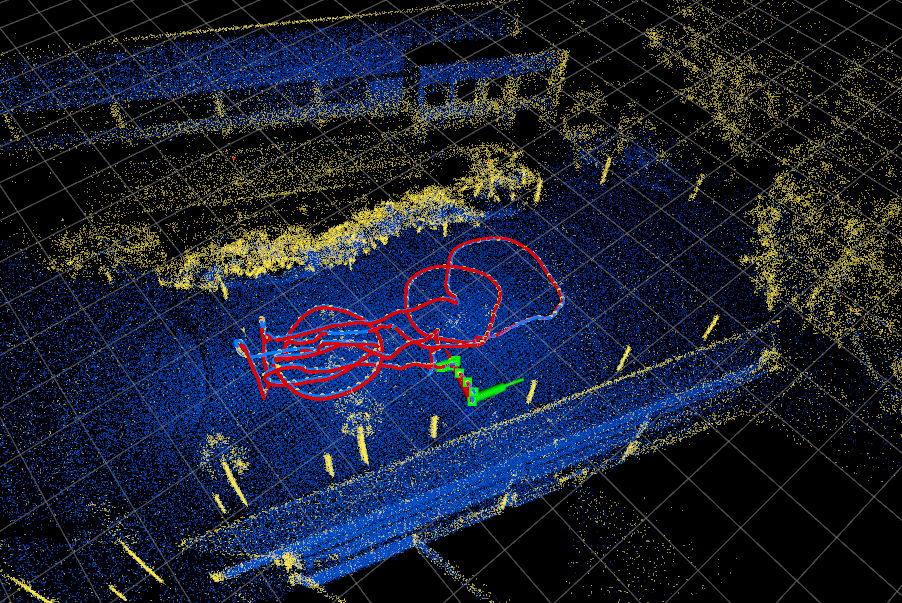}
    \caption{Visualization of the trajectory estimate by MILIOM method (blue line), ground truth (red line), key frame positions (yellow circles), activated key frames used for constructing the local map (light green), and the global map built by merging all of the key pointclouds. This experiment is done over the sbs\_03 dataset.}
    \label{fig: traj map kf}
\end{figure}

It is clearly shown in Tab. \ref{tab: ATE} that our system consistently achieves better performance compared to existing methods, as MILIOM with a single lidar already performs better than other methods, \highlight{including multi-lidar MLOAM method}. When configured to work with both lidars, the result is also consistently improved compared to the single lidar case.
In Fig. \ref{fig: eee 02 err}, we can observe a maximum error of 10cm by MILIOM method with two lidars. We also visualize our mapping and key frame management schemes in Fig. \ref{fig: traj map kf}.


\begin{figure}[h]
    \centering
    \includegraphics[width=\linewidth]{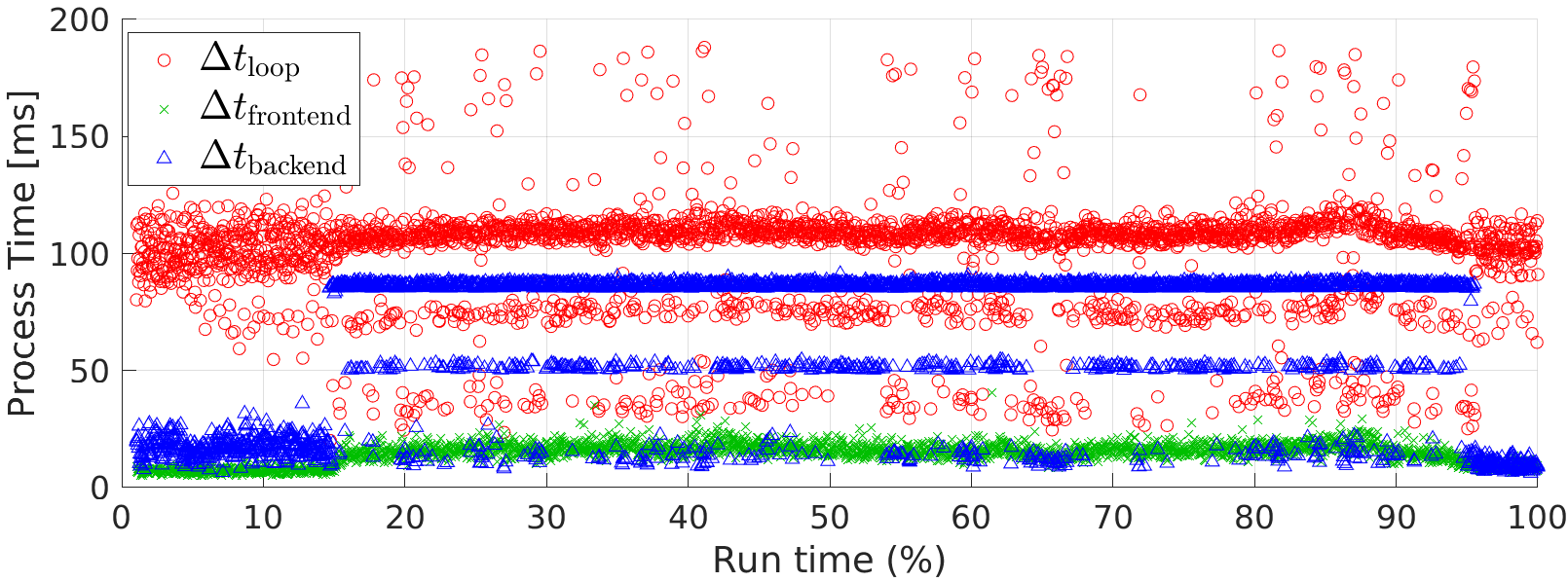}
    \caption{\highlight{Computation times of the main MILIOM processes in eee\_02 dataset: $\Dt_\text{loop}$ is the time period between the beginning of one joint optimization process to the next, $\Dt_\text{backend}$ is the time for the ceres-solver to actually solve the joint optimization problem \eqref{eq: cost function}, $\Dt_\text{frontend}$ is the computation time for all of the preliminary processes including deskew, transform and FMM processes before the optimization starts.}}
    \label{fig: solving times}
\end{figure}

\highlight{Fig. \ref{fig: solving times} reports the computation time for the main processes when running the eee\_02 dataset. The frontend processing time, which is mainly consumed by the FMM process, has a mean of 14.56 ms. While the backend processing time, dominated by the optimization time of the ceres solver, has a mean of about 62.9902 ms. The mean of the loop time is 99.99 ms, which exactly matches the lidar rate. The fluctuation in $\Dt_\text{loop}$ is reflective of the soft synchronization scheme when sometimes we have to wait for some extra time for data from all lidars to arrive, and sometimes we can immediately start on new frontend processing task when the current backend completes just when all lidar inputs have arrived at the buffer.}

\subsection{Building Inspection Trials}

\begin{table}[h]
\centering
\caption{\highlight{ATE of the lidar-based localization methods on some object inspection trials. The best result is highlighted in \textbf{bold}, the second best is \ul{underlined}. All values are in m.} \label{tab: bca trials}}
\renewcommand{\arraystretch}{1.1}
\begin{tabular}{c|ccc|cc}
\hline\hline
\mc{1}{c|}{\bf{Dataset}}
        & \mc{3}{c|}{\bf 1-Lidar}
        & \mc{2}{c}{\bf 2-Lidar}\\\cline{2-6}
        & \bf{LIO-M} & \bf{LIO-SAM} &\bf{Ours} &\bf{MLOAM} &\bf{Ours}\\\hline
test\_01 &3.6611     & -    & \ul{2.2460}    &3.9609      &\bf{0.2524}   \\
test\_02 &5.0644     & -    & 2.0184         &\ul{0.6699} &\bf{0.3378}   \\
test\_03 &5.7629     & -    & 4.4470         &\ul{3.4921} &\bf{0.1444}   \\\hline\hline
\end{tabular}
\end{table}
\begin{figure}[h]
    \centering
    \includegraphics[width=\linewidth]{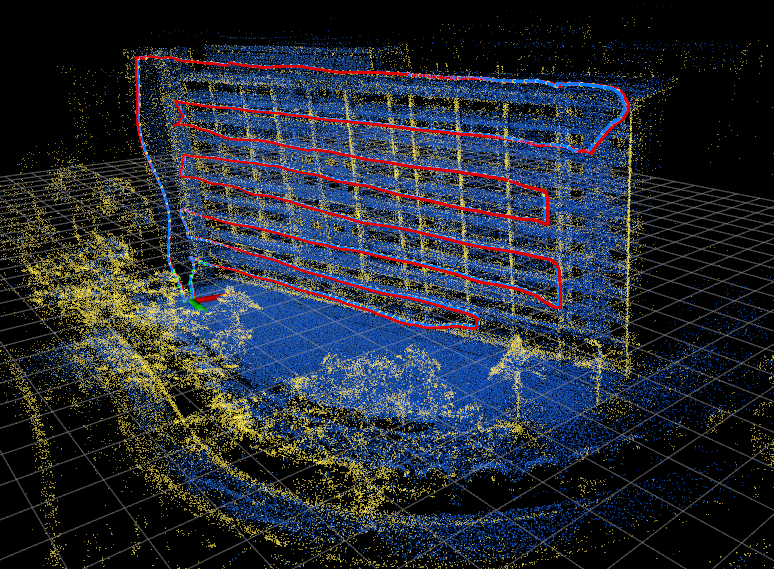}
    \caption{Visualization of the result in on the building inspection test\_01.}
    \label{fig: blk321 full}
\end{figure}

We also conduct several trials to test the ability of the method in actual inspection scenarios, one of which was shown earlier in Fig. \ref{fig: aerial loam challenge}. Indeed, these operations can be more challenging than the NTU VIRAL datasets for several reasons. First, there are only a building facade and the ground plane where reliable features can be extracted. Second, the building is much taller than the structures in the NTU VIRAL datasets, and when flying up to that high, there might be very few features that can be extracted. Finally, the trajectory of the UAV is mostly confined to the YZ plane, instead of having diverse motions as in the NTU VIRAL datasets. Hence there can be some observability issues with the estimation.

Tab. \ref{tab: bca trials} summarizes the results of these inspection trials. In general, the aforementioned challenges do increase the error for all methods. We couldn't get LIO-SAM to work in these datasets till the end without divergence. Even for the MILIOM method with a single lidar, there is also a significant increase in the error. \highlight{On the other hand, contrary to its lower performance over NTU VIRAL datasets, in these challenging building inspection trials, MLOAM can take advantage of complementary FOVs of both lidar, thus achieving significantly higher accuracy than single-lidar MILIOM. Nevertheless, the results of the two-lidar MILIOM approach are still most accurate, which demonstrates the robustness and accuracy of the multi-input lidar-inertia} approach in critical conditions. Fig. \ref{fig: blk321 full} presents the 3D plot of the trajectory estimates, the global map and groundtruth in one of these datasets.

\section{Conclusion and Future Works} \label{sec: conclusion}

In this paper we have developed a tightly-coupled, multi-threaded, multi-input, key frame based LIOM framework. We demonstrated our system's capability to estimate the position of a UAV and compared the performance with other state of the art methods. The results show that our method can outperform existing methods in single lidar case, and the use of multiple lidars can further improve the accuracy as well as the robustness of the localization process. We shown that our method can guarantee real time processing capability.

In the future, we will improve this localization system by combining it with other types of sensor such as visual features and ranging measurements. In addition, loop closure and bundle adjustment are also being investigated.

\balance
\bibliographystyle{IEEEtran}
\bibliography{references, mypublications}

\end{document}